\documentclass[11pt,a4paper]{article}
\usepackage[hyperref]{acl2019}
\usepackage{times}
\usepackage{latexsym}

\usepackage{url}
\usepackage{graphicx}
\usepackage{algorithm}
\usepackage{algorithmic}
\usepackage{soul}
\usepackage{amssymb}
\usepackage{contour}
\usepackage{booktabs}
\usepackage{multirow}
\usepackage{siunitx}

\aclfinalcopy 

\title{Specificity-Based Sentence Ordering \\for Multi-Document Extractive Risk Summarization}

\author{Berk Ekmekci, Eleanor Hagerman and Blake Howald \\
	Thomson Reuters Special Services, LLC\\
	1410 Spring Hill Road, Suite 301 \\
	Mclean, VA 22102-3058\\
	{\tt firstname.lastname@trssllc.com}  }
\date{}

\begin{document}
\maketitle
\begin{abstract}
Risk mining technologies seek to find relevant textual extractions that capture entity-risk relationships. However, when high volume data sets are processed, a multitude of relevant extractions can be returned, shifting the focus to how best to present the results. We provide the details of a risk mining multi-document extractive summarization system that produces high quality output by modeling shifts in specificity that are characteristic of well-formed discourses. In particular, we propose a novel selection algorithm that alternates between extracts based on human curated or expanded autoencoded key terms, which exhibit greater specificity or generality as it relates to an entity-risk relationship. Through this extract ordering, and without the need for more complex discourse-aware NLP, we induce felicitous shifts in specificity in the alternating summaries that outperform non-alternating summaries on automatic ROUGE and BLEU scores, and manual understandability and preferences evaluations - achieving no statistically significant difference when compared to human authored summaries. 
\end{abstract}

\section{Introduction}

\textit{Risk mining} seeks to identify the expression of entity-risk relationships in textual data \cite{LeidnerSchilder2010}. For example, (1a-b) describe a \textit{CNN-Terrorism} relationship that is indicated by the reference to \textbf{CNN} in (1a) and \textbf{Terrorism} risk category keywords - \underline{pipe bomb} (1a) and \underline{bomb threat} (1b).\\

\footnotesize 
\noindent(1a) \textit{Later Wednesday, \textbf{CNN} received a \underline{pipe bomb} at its Time Warner Center headquarters in Manhattan sent to ex-CIA director John Brennan,prompting \textbf{CNN} to evacuate its offices.}\\ 

\noindent(1b) \textit{It was the second time in two days that the building was evacuated in a \underline{bomb threat}.}

\enspace
\normalsize
The goal of risk mining systems is to identify the highest value and most relevant text extractions that embody an entity-risk relationship, indexed by an entity and a keyword/phrase - obviating the need for a manual review of numerous sources. However, as systems expand both in volume of data analyzed and discovery of new entity-risk relation expressions, the number of relevant extracts increases and the challenge to review the information returns. We rely on extractive summarization (\textit{see generally}, \citet{NenkovaMcKeown2011}) to address this problem with particular emphasis on creating high quality output that appropriately orders extracted clauses by information specificity.

To illustrate, (1a) provides specific referents about time (\textit{Later Wednesday}), events (\textit{receiv[ing] a pipe bomb}), location (\textit{Time Warner Center headquarters in Manhattan}), people (\textit{ex-CIA director John Brennan}), and the resulting event of \textit{evacuat[ing] its [CNN's]} offices. (1b), through the use of less specific references (\textit{It, the building}) and its sequencing relative to (1a), generalizes that this was the second such event in two days. If we reorder (1a-b) in (2a-b), which can happen in extractive systems, the flow of information is less felicitous and reads less easily.  

\enspace
\footnotesize
\noindent(2a) \textit{It was the second time in two days that the building was evacuated in a \underline{bomb threat}.}\\ 

\noindent(2b) \textit{Later Wednesday, \textbf{CNN} received a \underline{pipe bomb} at its Time Warner Center headquarters in Manhattan sent to ex-CIA director John Brennan, prompting \textbf{CNN} to evacuate its offices.}

\enspace
\normalsize
One way to improve output in these circumstances is to control sentence ordering. This is simpler in single documents as preserving the order of the extract in the documents works to encourage a coherent summary  (e.g., \citet{133McKeown:1999:TMS:315149.315355}). However, for multi-document summaries, this is not as simple and approaches to sentence ordering become much more complex.

We propose a novel approach to ordering extractive summaries by focusing on the specific or general nature of the extracts when building the summary. In particular, we identity two groups of extracts from a keyword-based risk mining system: one characterized as more specific (from a manually curated set of keywords) and one characterized as more general (from a semantically encoded set of keywords). Alternating the extract selection between these two groups, which are ranked by bidirectional token distances between the entity and the risk keyword, creates extractive summaries that outperform non-alternating systems - so much so that our top performing system fails to be significantly different from the comparative human authored summaries.

In this paper, we review risk mining, extractive summarization, and discuss information specificity in discourse (Section 2). Section 3 presents our risk mining system, with emphasis on: entity-keyword extraction, the expansion of the human curated taxonomy, and the nature of these extraction sets relative to specificity in discourse. Section 4 introduces and presents the results of several experiments evaluated with automatic ROUGE, BLEU, and manual preference and readability judgments. We discuss the results and related work in extractive summarization (Section 5), and discuss future work in Section 6.

\section{Background}

This section provides a high-level overview of risk mining (Section 2.1), automatic summarization (Section 2.2.), and the relationship between specificity and discourse (Section 2.3) to contextualize the presentation of our system in Section 3. Citations are not meant to be exhaustive. Additional treatment of automatic summarization comparable to ours is discussed in Section 5.

\subsection{Risk Mining}
Risk mining systems typically start with a keyword list that captures, from a subject matter expert's perspective, a risk category of interest and entities that are subject to that risk (e.g., media outlets subject to terrorism, persons subject to fraud). Systems also expand this ``seed" keyword list and fine tune output through some combination of machine learning and human-in-the-loop review until a desired level of performance is achieved \cite{LeidnerSchilder2010, NugentLeidner2017}. Domains where risk mining has been applied include financial risks based on filings and stock prices \cite{KoganEtAl2009,DasguptaEtAl2016}; general risks in news \cite{LuEtAl2009b,NugentEtAl2017}, and supply chain risks \cite{CarstensEtAl2017}. Further, methods of keyword list expansion include ontology merging \cite{SubramaniamEtAl2010}, crowdsourcing \cite{MengEtAl2015} and paraphrase detection \cite{PlachourasEtAl2018}. The goal of the expansion is to minimize human involvement while still preserving expert judgment, maintaining and improving performance through the return of highly relevant extracts. 

\subsection{Automatic Summarization}
Approaches to automatic text summarization fall into either the \textit{abstractive} or \textit{extractive} categories. Abstractive approaches seek to identify relevant phrases and sentences. The summary is a rewriting of those extracts; with recent approaches making use of graphs \cite{tan-etal-2017-abstractive, dohare-etal-2018-unsupervised} or neural networks  \cite{ chopra-etal-2016-abstractive,Paulus2018ADR}. Extractive approaches attempt to: identify relevant text extractions in single and multi-document source material; rank the extracts to find the most informative; and combine the selected extracts into a summarized discourse. 

Finding and ranking relevant extracts is based on queries \cite{7377323}, document word frequencies \cite{40conroy-etal-2006-topic,70Gupta:2007:MIQ:1557769.1557825}, probabilities \cite{212Vanderwende:2007:BST:1284916.1285165}, tf-idf weighting \cite{[56]Erkan:2004:LGL:1622487.1622501,61Fung:2006:OSO:1149290.1151099}, topic modeling \cite{106Lin:2000:AAT:990820.990892}, sentence clustering \cite{133McKeown:1999:TMS:315149.315355,190siddharthan-etal-2004-syntactic}, graph-based methods \cite{55Erkan04theuniversity,[56]Erkan:2004:LGL:1622487.1622501,mihalcea-tarau-2004-textrank}, and neural networks \cite{filippova-etal-2015-sentence,DBLP:conf/aaai/NallapatiZZ17}. Our extraction method (Section 3.2) is based on entity-keyword matching in multiple documents with subsequent ranking of token distances between entities and risk keywords. 

Once extracts are selected for inclusion, techniques are applied to improve the overall quality of the summary. Improvements on the sentence level include sentence compression \cite{206turner-charniak-2005-supervised,65galley-mckeown-2007-lexicalized} and fusion \cite{90jing-mckeown-2000-cut,11barzilay-mckeown-2005-sentence}. Improvements on the semantic and pragmatic level include use of lexical chains \cite{7Barzilay97usinglexical,64Galley:2003:IWS:1630659.1630906}, WordNet \cite{Fellbaum98} -based concepts \cite{186Schiffman:2002:EMS:1289189.1289254}, Latent Semantic Analysis \cite{69Gong:2001:GTS:383952.383955,73Hachey:2006:DRA:1654679.1654681}, and discourse relation and graph representations \cite{159ono-etal-1994-abstract,120Marcu1998ToBT,119marcu-1997-discourse,122Marcu:2000:TPD:517637,Wang:2015:SBT:2876444.2876454}. As discussed in Section 5, our system most closely aligns with \textit{sentence ordering} methods of improvement in multi-document extractive summarization research.

\subsection{Specificity and Discourse Structure}
At the word level, specificity can be defined in terms of \textit{generics} and \textit{habituals} in (3a-d):

\enspace
\footnotesize
\noindent(3a) \textbf{Generic:} 

\hspace{5mm}\textit{Dogs love to go for walks.} \\
\noindent(3b) \textbf{Non-Generic:} 

\hspace{5mm}\textit{The dog is in the backyard.}\\
\noindent(3c) \textbf{Habitual:} 

\hspace{5mm}\textit{She had trouble walking after she slipped and fell.}\\
\noindent(3d) \textbf{Non-Habitual:} 

\hspace{5mm}\textit{She slipped and fell in January of 2019.}

\enspace
\normalsize
\noindent \textit{Generics} describe either a class of entities - \textit{dogs} in (3a), or a member of a class of entities - \textit{the dog} in (3b). \textit{Habituals} describe either specific or regular events - \textit{trouble walking} (3c) - \textit{slipped and fell} (3d). The ability to detect generics and habituals computationally relies on word-level features such as plurals, quantifiers, verb tenses, categories of noun phrases, and lexical resources such as WordNet (\textit{see generally,} \citet{MathewKatz2009, Friedrich2015DiscoursesensitiveAI}).

Beyond the sentence, \citet{Li2017} links occurrences to information specificity to rhetorical relations. For example, the \textsc{background} relation provides general backdrop information for subsequent clauses; \textsc{elaboration} provides more specific unfolding of events; and \textsc{specification} provides more specific detail of the previous information (\textit{see e.g.}, Rhetorical Structure Theory \cite{MT:87} and the Penn Discourse TreeBank \cite{P:08}). \citet{Mulkar-Mehta:2011:GNL:2002669.2002712} weave generics and habituals into a ``granularity" framework of \textit{part-of} and causality shifts across clauses in discourse. \citet{DBLP:conf/starsem/HowaldA12} and \citet{DBLP:conf/cosit/HowaldK11} demonstrate that annotated granularities improved machine learning prediction of Segmented Discourse Representation Theory \cite{AL:03} rhetorical relations.  

Appropriately ordered shifts in specificity are generally associated with texts of higher quality \cite{DBLP:conf/ijcnlp/LouisN11A}, which can be interpreted as increased readability \cite{Dixon82, Dixon87}, higher coherence \cite{Hobbs85,Kehler02} and accommodation of the intended audience \cite{BC08,Djalalietal2011}. \citet{DBLP:conf/acl/LouisN11B} further observe that automatic summaries tend to be much more specific than their human authored counterparts and are judged to be incoherent and of lower comparative quality. As discussed in Section 4, rather than explicitly identifying and exploiting habituals, generics or rhetorical relations, we model shifts in specificity by alternating selection from sets of extracts that are characterized as more or less specific relative to an entity-risk relation as a byproduct of risk term expansion.

\section{System}

Our initial extraction system is a custom NLP processing pipeline capable of ingesting and analyzing hundreds of thousands of text documents relative to a manually-curated seed taxonomy. The system consists of five components:

\footnotesize
\begin{enumerate}
	\item \textbf{Document Ingest and Processing:} Raw text documents are read from disk and tokenization, lemmatization, and sentencization are performed. 
	\item \textbf{Keyword/Entity Detection:} Instances of both keywords and entities are identified in the processed text, and each risk keyword occurrence is matched to the nearest entity token.
	\item \textbf{Match Filtering and Sentence Retrieval:} Matches within the documents are filtered, categorized by pair distance, and corresponding spans retrieved for context. For comparison to methods relying on sentence co-occurrence, the sentences are retrieved for context.
	\item \textbf{Semantic Encoding and Taxonomy Expansion:} A semantic vectorization algorithm is trained on domain-specific texts and used to perform automated expansion of the keyword taxonomy.
	\item \textbf{Extractive Summarization Construction:} From the total collection of extracts, summaries are formed based on different combinations distances, keyword frequencies, and taxonomy.  
\end{enumerate}

\normalsize

We leverage \textit{spaCy} (Version 2.0.16, https://spacy.io) \cite{spacy2} as the document ingest and low-level NLP platform for this system. This choice was influenced by spaCy's high speed parsing \cite{ChoiEtAl2015}, out-of-the-box parallel processing, and Python compatibility. In particular, spaCy's \textit{pipe()} function allows for a text generator object to be provided and takes advantage of multi-core processing to parallelize batching. In this implementation, each processed document piped in by spaCy is converted to its lemmatized form with sentence breaks noted so that sentence and multi-sentence identification of keyword/entity distances can be captured.

\subsection{Keyword/Entity Detection}
The shorter the token distance between entity and keyword, the stronger the entity-risk relationship is as a function of semantic and pragmatic coherence. (4) describes the entity \textbf{Verizon} and its litigation risk associated with lawsuit settlement (indicated by \underline{settle} and \underline{lawsuit} keywords).

\enspace
\footnotesize
\noindent(4) \textit{In 2011, \textbf{Verizon} agreed to pay \$20 million to \underline{settle} a class-action \underline{lawsuit} by the federal Equal Employment Opportunity Commission alleging that the company violated the Americans with Disabilities Act by denying reasonable accommodations for hundreds of employees with disabilities.}

\enspace
\normalsize

We return the entire sentence to provide additional context - a \textit{class-action} lawsuit and the allegation that \textbf{Verizon} \textit{denied reasonable accommodations for hundreds of employees with disabilities.} Extracts can further improve when the distances are considered bidirectionally. For example, (5) extends (4) to the prior contiguous sentence which contains \textit{settlement}. This extension provides greater context for Verizon's lawsuit. (5) contains a \textsc{background} relation and provides the larger context that Verizon is in violation of settlement terms from a previous lawsuit.

\enspace
\footnotesize
\noindent(5) \textit{McDonald says this treatment violated the terms of a \underline{settlement} the company reached a few years earlier regarding its treatment of employees with disabilities. In 2011, \textbf{Verizon} agreed to pay \$20 million to settle a class-action lawsuit by the federal Equal Employment Opportunity Commission ....}

\enspace
\normalsize

The system detection process begins by testing for matches of each keyword with each entity, for every possible keyword-entity pairing in the document. For every instance of every keyword, the nearest instance of every available entity is paired regardless of whether it comes before or after the keyword (\textbf{Algorithm 1}). An entity may be found to have multiple risk terms associated with it, but each \textit{instance} of a risk term will only apply itself to the closest entity - helping to minimize overreaching conclusions of risk while maintaining system flexibility. 

\begin{algorithm}[h]
	\footnotesize
	\caption{Entity-Keyword Pairing}
	\begin{algorithmic}
		\REQUIRE taxonomy and entities lists 
		\FOR{keyword in taxonomy}
		\FOR{entity in entities}
		\STATE keywordLocs = findLocs(keyword)
		\STATE entityLocs = findLocs(entity)
		\FOR{kLoc in keywordLocs}
		\STATE bestHit = findClosestPair(kLoc, entityLocs)
		\STATE results.append((keyword, entity, bestHit))
		\ENDFOR
		\ENDFOR
		\ENDFOR
		\RETURN findClosestPair is two token indicies
	\end{algorithmic}
\end{algorithm}

The system's token distance approach promotes efficiency compared to more complex NLP. However, the computational cost of this is: a total of $ (i \cdot a) \times (j \cdot b) $ comparisons must be made for each document, where \textit{i} is the number of keyword terms across all taxonomic categories, \textit{a} the average number of instances of each keyword per document, \textit{j} the number of entities provided, and \textit{b} the average number of entity instances per document. Changing any single one of these variables will result in computational load changing with $\mathcal{O}(n)$ complexity, but their cumulative effects can quickly add up.\footnote{For parallelization purposes, each keyword and entity is independent of each other keyword and entity. This means that in an infinitely parallel (theoretical) computational scheme, the system runs on $\mathcal{O}(a \times b)$, which will vary as a function of the risk and text domains.} 

\begin{table*}[t]
	\small
	\centering
	\begin{tabular}{|p{1.75cm}|p{6.25cm}|p{6.25cm}|}
		\hline
		\textbf{Category}&\textbf{Seed} & \textbf{Expanded}\\
		\hline
		\textit{Cybersecurity} & \textit{n=20} & \textit{n=32 (additional)}\\
		\hline
		&\textit{antivirus, cybersecurity, data breach, denial of service, hacker, malware, network intrusion, phishing, ransomware, spyware, virus, ...} & \textit{\textbf{4frontsecurity}, \textsc{attack}, \textsc{beware}, \textbf{cyberattack}, \textbf{cyberstalking},\textsc{detection}, \textsc{identity}, \textbf{opsware}, \textbf{phish}, \textsc{ransom}, \textsc{security}, \textbf{socialware}...}\\
		
		\hline
		\textit{Terrorism} & \textit{n=23}& \textit{n=47 (additional)}\\
		\hline
		&\textit{bioterrorism, car bomb, counterterrorism, extremist, hijack, jihad, lone wolf, mass shooting, separatism, suicide bomber, terrorist, ...}& \textit{\textbf{bombmaker}, \textsc{consequence, criticism}, \textbf{fascist}, \textsc{hate}, \textbf{hezbollah}, \textbf{hijacker}, \textbf{jihadi}, \textsc{massive, military, suspicious}, ...}\\
		
		\hline
		\textit{Legal} & \textit{n=26}& \textit{n=54 (additional)}\\
		\hline
		&\textit{allegation, bankruptcy, indictment, infringement, lawsuit, litigation, misappropriation, negligence, plaintiff, regulatory violation, statutory, ...}&\textit{\textsc{action, carelessness}, \textbf{extortion, foreclosure, infringe, pre-litigation}, \textsc{reckless}, \textbf{relitigate}, \textsc{require, suit}, \textbf{tort}, ...}\\
		\hline
	\end{tabular}
	\caption{Sample risk terms: qualitatively \textbf{bolded} terms are more specific and \textsc{smallcaps} terms are more general relative to the risk category.}
\end{table*}

\subsection{Encoding}
Our system automates term expansion by using similarity calculations of semantic vectors. These vectors were generated by training a \textit{fastText} (https://fasttext.cc/) skipgram model \cite{BojanowskiEtAl2017}, which relies on words and subwords from the same data sources identified in the initial run of the system using the seed taxonomy. This ensures that domain usage of language is well-represented, and any rich domain-specific text may be used to train semantic vectors (\textit{see generally}, \citet{MikolovEtAl2013}). 

For each risk term, the system searches the model vocabulary for the minimized normalized dot product $\frac{r \cdot w}{\| r \|\| w \|}$ (a basic similarity score found in the \textit{fastText} codebase), and returns the top-scoring vocabulary terms. Upon qualitative review, the expansion finds new keywords that are specific to the entity-risk relationship, but a higher proportition of new keywords that are more general (Table 1).\footnote{The method also produces tokenized variants and misspellings (\textit{neg igence, thisagreement}), items clearly out of semantic bounds (\textit{gorilla, papilloma, titration}), and substring drifting (\textit{fines,vines,wines}). These are low frequency and typically down selected by the system rather than removed).}
\begin{table}[h]
	\centering
	\begin{tabular}{|l|c|c|c|c|}
		\hline
		\textbf{Category}& \textbf{Seed} & \textbf{Expanded} &\textbf{\% $\uparrow$}\\
		\hline
		\textit{Cybersecurity} & 1.40 & 2.41 & 72.14\\
		\hline
		\textit{Terrorism} & 2.13 & 2.46 & 15.49\\
		\hline
		\textit{Legal} & 1.73 & 3.60 & 108.09\\
		\hline
	\end{tabular}
	\caption{WordNet Polysemy Seed and Increased Expanded Averages.}
\end{table}

To illustrate more quantitatively, the possibility of the content of the extracts having a general or specific character is indicated in Table 2. We calculated a polysemy average: for every word in the keyword sets, we averaged the number of definitions per word from WordNet. The higher the number (the more definitions) the more general the keyword can become relative to the context. Greater increases are seen for \textit{Cybersecurity} and \textit{Legal} (more general) compared to \textit{Terrorism} where the expansion appears to have maintained a similar mix of specific and general. While the filtering of documents by entities may somewhat control the contexts, there is, of course, no guarantee of this. However, we suggest that our method benefits from operating within a specified entity-risk relationship (controlling the extraction, expansion and source material).  

\subsection{Selection}
After processing, the resulting extracts are deduped, preserving the lowest distance version. Remaning extracts are ranked by highest frequency keyword and then by shortest distance within the keyword. Summary extract selection proceeds as follows (\textbf{Algorithm 2}):

\begin{algorithm}[h]
	\footnotesize
	\caption{Extract Selection}
	\begin{algorithmic}
		\REQUIRE ranked list by distance and keyword frequency 
		\WHILE{summary is less than n number of words}
		\IF{keyword not in selectedWords}
		\STATE {summary+=top extract}
		\STATE {selectedWords+=keyword}
		\STATE {remove extract}
		\ELSE
		\STATE{rerank remaining results}
		\STATE{selectedWords=[]}
		\ENDIF
		\ENDWHILE
		\RETURN {summary}
	\end{algorithmic}
\end{algorithm}

For experimentation (Section 4), we first selected the top Fortune 100 companies from 2017 (http://fortune.com/fortune500/2017/) as input (\textit{entities}) into a proprietary news retrieval system for the most recent 1,000 articles mentioning each company (\textit{sources}). Ignoring low coverage and bodiless articles, 99,424 individual documents were returned. Second, each article was fed into the system and risk detections for three risk relationships (\textit{Cybersecurity, Terrorism}, and \textit{Legal}) were found with a distance cutoff of 100 (word) tokens. Lastly, a baseline extract was selected at random for each identified risk from the corresponding document for pairwise comparison. The probability of a multi-sentence extract occurring in the output is high - approx. 70\% with an average token distance of 30 for multi- or single sentence extraction (standard deviation is as high as 25 tokens).  

\begin{table*}[t]
	\small
	\centering
	\begin{tabular}{|c|p{13cm}|}
		\hline
		\textbf{System}&\textbf{Costco-Legal}\\
		\hline
		\textit{\textit{Human}} & \textit{A lawsuit was brought against  Costco for negligence, carlessness, and having defective conditions. Costco is also being investigated for potential corporate misconduct  concerning sales of products that are alleged to be counterfeit and/or to Infringe patent/trademark rights. The Acushnet Company who is the holder of certain Titleist golf ball patnets is also in litigation with Costco alleging patent infringement and false advertising.}\\
		\hline
		\textit{Alternate Thirds} & \textit{The suit claims Costco should be held liable for the injuries due to its "negligence and carelessness," and for having "dangerous or defective conditions." In addition to the litigation with Tiffany \& Co., the Company has also recently been in litigation with Acushnet Company, represented to be the holder of certain Titleist golf ball patents, concerning allegations that Costco has committed patent infringement and false advertising in connection with the sale of certain golf balls in its membership warehouses. The plaintiffs did not accept Costcos proposals for settlement and Costco ultimately prevailed on a motion for summary judgment.}\\
		\hline
		\textit{Mixed Thirds} & \textit{The suit claims Costco should be held liable for the injuries due to its "negligence and carelessness," and for having "dangerous or defective conditions." In her motion, Pronzini challenges Costcos allegation that it is headquartered in Washington. The lawsuit claims Costco should have known about any "unsafe, dangerous or defective conditions" in the store.} \\
		\hline
		\textit{Expansion} & \textit{Costco's settlement of this matter does not constitute an admission of staff's charges as set forth in paragraphs 4 through 12 above. In addition to the litigation with Tiffany \& Co., the Company has also recently been in litigation with Acushnet Company, represented to be the holder of certain Titleist golf ball patents, concerning allegations that Costco has committed patent infringement and false advertising in connection with the sale of certain golf balls in its membership warehouses.}\\	
		\hline
	\end{tabular}
	\caption{Sample Expanded and Human Summaries for Costo-Legal Entity Risk Relationship.}
\end{table*}

\section{Evaluation and Results}
We evaluate four systems that produce different combinations of general and specific extracts:

\footnotesize
\begin{itemize}
	\item \textit{Seed} - Seed extracts only.
	\item \textit{Expanded} - Expanded extracts only.
	\item \textit{MixedThirds} - The first selection is from the expanded set, all remaining selections are from the seed set.
	\item \textit{AlternateThirds} - Selection proceeds from expanded to seed to expanded.
\end{itemize}
\normalsize
Depending on the specificity or generality of a given clause,  a pair of extracts may flow from general to specific or vice versa. We choose a canonical narrative flow for the overall text - i.e., general to specific (and back to general) (\textit{see e.g.} \citet{Labov704}) - which is tested in the \textit{MixedThirds} and \textit{AlternateThirds} systems. Table 3 provides example output for the \textbf{Costco-Legal} entity-risk relation, thresholded to 100 words.

We further test a random baseline system as well as two existing extractive summarization systems \textit{TextRank} \cite{mihalcea-tarau-2004-textrank} and \textit{LexRank} \cite{Radev04lexrank:graph-based}.\footnote{\textit{TextRank} implemented with Summa NLP's Textrank - https://github.com/summanlp/textrank and \textit{LexRank} implemented with Crabcamp's Lexrank - https://github.com/crabcamp/lexrank.}

\footnotesize
\begin{itemize}
	\item \textit{Baseline} - For a given entity risk relationship, extracts are randomly selected until the 100 word limit is reached.
	\item \textit{TextRank} - Each extract is a node in a graph with weighted edges by normalized word overlap between sentences.
	\item \textit{LexRank} - Each extract is a node in a graph with weighted edges based on cosine similarity of the extract set's TF-IDF vectors.
\end{itemize}
\normalsize

We asked six analysts (subject matter experts in risk analysis) to write human summaries for each entity-risk relationship relying on reference extracts filtered by lowest distance and keyword. These human summaries, also thresholded at 100 words, were used in `intrinsic' comparison evaluations (how informative the summaries are) with ROUGE - Recall-Oriented Understudy for Gisting Evaluation \cite{lin:2004:ACLsummarization} and BLEU (Bilingual Evaluation Understudy) \cite{Papineni:2002:BMA:1073083.1073135} \cite{Loper:2002:NNL:1118108.1118117}.\footnote{ROUGE implemented with Kavita Ganesan's JAVA Rouge 2.0 - https://github.com/kavgan/ROUGE-2.0 and BLEU implemented with the Natural Language Tool Kit (NLTK).} ROUGE and BLEU alone can be limited without additional `extrinsic' evaluations (how well the summaries are formed) to support and appropriately characterize results. Consequently, we conducted two additional manual evaluations: an \textbf{A/B Preference Judgment} task, pitting all systems against human summaries, and a \textbf{Readability Judgment} task using a 3-Point scale: \textit{Fluent} (5) = no grammatical or informative barriers, \textit{Understandable} (3) = some grammatical or informative barriers, \textit{Disfluent} (1) = significant grammatical or informative barriers. 

\begin{table*}[t]
	\centering
	\begin{tabular}{|l|c|c|c|c||c|}
		\hline
		\textbf{System} & ROUGE-1 & ROUGE-2 & ROUGE-L & ROUGE-SU & BLEU-4\\
		\hline
		\hline
		\textit{Seed} & 9.18 & 2.78 & 8.04 & 3.45 & 29.48\\
		\hline
		\textit{Expanded} & \textbf{20.45} & \textbf{10.81} & \textbf{18.35} & \textbf{11.55} & 30.22\\
		\hline
		\textit{MixedThirds} & \textbf{12.29} & \textbf{4.11} & \textbf{10.43} & \textbf{4.93} & \textbf{31.79}\\
		\hline
		\textit{AlternateThirds} & \textbf{18.12} & \textbf{8.51} & \textbf{15.66} & \textbf{9.37} & \textbf{32.05}\\
		\hline
		\textit{Baseline} & 9.74 & 3.35 & 9.33 & 4.03 & \textbf{30.61}\\
		\hline
		\textit{TextRank} & 8.05 & 2.80 & 9.01 & 3.24 & 28.62\\
		\hline
		\textit{LexRank} & 9.48 & 2.83 & 8.74 & 3.53& 29.96\\
		\hline
	\end{tabular}
	\caption{ROUGE-1,-2,-L,-SU Average $F_{1}$ and BLEU-4 Results (top three scores \textbf{bolded}).}
\end{table*}

\subsection{Results}
We focus results on the system level for simplification as performance was similar across all risk categories and evaluations. In Table 4, we report average $F_{1}$ for unigram (ROUGE-1), bigram (ROUGE-2), longest common subsequence (ROUGE-L), and skip-4-gram using unigram co-occurrence statistics (ROUGE-SU) and the BLEU (4-gram) score. Each system summary was compared against two human summaries from the same entity-risk relationship. System summaries that pulled from the expanded (more general) set of extractions performed best across all versions of ROUGE and BLEU-4, with \textit{MixedThirds} and \textit{AlternativeThirds} outperforming all other systems. 

\begin{table}[h]
	\centering
	\begin{tabular}{|l|c|}
		\hline
		\textbf{System} &\textbf{$\chi^2$} (\textit{p} [d.f.=1])\\
		\hline
		\hline
		\textit{Seed} & 17.64 (\textit{p}$<$0.001) \\
		\hline
		\textit{Expanded} & 12.82 (\textit{p}$<$0.001)\\
		\hline
		\textit{MixedThirds} & 11.68 (\textit{p}$<$0.001)\\
		\hline
		\textbf{\textit{AlternateThirds}} & \textbf{3.68 (\textit{p}$>$0.05)}\\
		\hline
		\textit{Baseline} & 23.12 (\textit{p}$<$0.001)\\
		\hline
		\textit{TextRank} & 49.08 (\textit{p}$<$0.001)\\
		\hline
		\textit{LexRank} & 13.62 (\textit{p}$<$0.001)\\
		\hline
	\end{tabular}
	\caption{Pearson's $\chi^2$ for Preference Judgments. No statistically significant difference when \textit{AlternateThirds} is compared to human summaries (\textit{p}$>$0.05).}
\end{table}

For A/B Preference Judgments, 2,000 annotations (1,000 double annotated instances) were collected for human summaries versus all systems. There is a trend of greater preference for the expanded over non-expanded systems (Figure 1). This is supported with Pearson's $\chi^2$ (Table 5) where there is no statistically significant difference between \textit{AlternateThirds} and human summaries. Statistically significant differences exist with and all other system comparisons, though a narrowing percentage preference gap for the expanded systems. Average $Kappa$ \cite{Cohen1960} for the Preference Judgment was quite low at .069, indicating not only the difficulty of the task and a significant source of disagreement among the risk analysts, but also increased randomization based on the lack of a third 'no difference' option.  

\begin{table}[h]
	\centering
	\begin{tabular}{|l|c|}
		\hline
		\textbf{System}& Readability\\
		\hline
		\hline
		\textit{Human} & 3.75\\
		\hline
		\textit{Baseline} & \textbf{2.54}\\
		\hline
		\textit{AlternateThirds} & \textbf{2.50}\\
		\hline
		\textit{Expanded} & \textbf{2.37}\\
		\hline
		\textit{Seed} & 2.31\\
		\hline
		\textit{MixedThirds} & 2.20\\
		\hline
		\textit{LexRank} & 2.14\\
		\hline
		\textit{TextRank} & 1.92\\
		\hline
	\end{tabular}
	\caption{Average Readability (1-3-5 Scale). \textit{AlternateThirds} and \textit{Baseline} (Discussed in Section 5) have the highest non-human readability across all systems.}
\end{table}
For Readability Judgments, 1,600 annotations were collected (800 doubly annotated instances) for all systems and human summaries. The human summaries garnered the highest scores with a 3.75 average (Table 6) with the \textit{Expanded} and \textit{AlternateThirds} (and \textit{Baseline}) achieving scores between 2.37 and 2.54. \textit{Alternate Thirds} and \textit{Expanded} also had the highest proportion of ``5'' ratings (20\%) compared to 50\% for the human summaries and 15\% or lower for the other systems. Average $Kappa$ improved to .163, but still low.
\enspace
\begin{figure*}[h]
	\includegraphics[width=\linewidth,height=6cm]{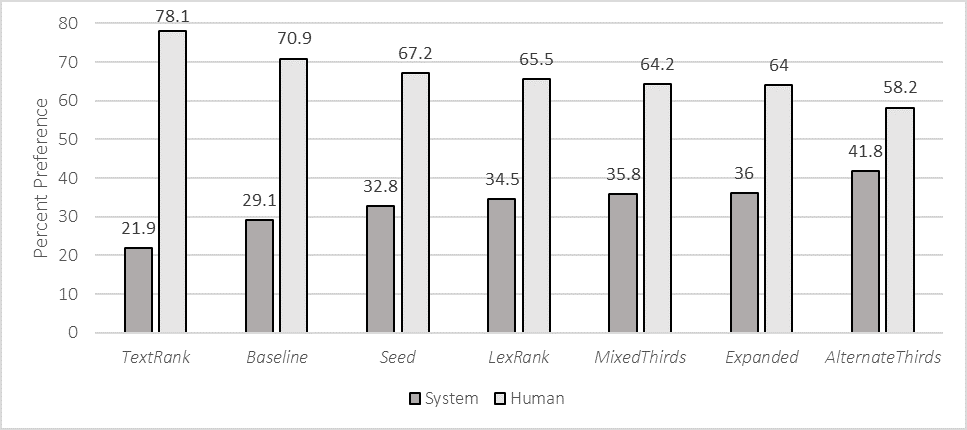}
	\caption{Expert preference ratings.}
\end{figure*}

\section{Discussion and Related Work}
Overall, \textit{Alternate} and \textit{MixedThirds} systems have the highest content overlap and are packaged in a way that yield high readability and preference ratings when compared to human summaries. When variation was observed in the results (low scores for these systems, or high scores for non-alternating systems) it often had to do with the experimental design rather than specificity ordering. For example, \textit{Baseline} extractions received ``5" ratings (c.f. Tables 4 and 6 for good \textit{Baseline} performance) when they were short coherent discourses (6):

\enspace
\footnotesize
\noindent(6)\textit{Well before a deranged anti-Semite opened fire in the Tree of Life Synagogue, instances of anti-Semitism and hate crimes were on the rise. White nationalists have felt emboldened to march in American cities. And days before the shooting, a gunman tried to shoot up a predominantly black church. When he failed, he went to a nearby Kroger outside Louisville, Kentucky, and killed two people there instead.}

\enspace
\normalsize 
\noindent Further, the performance of \textit{TextRank} and \textit{Lexrank} was likely inhibited by being run on the extracts rather than the documents themselves; though \textit{LexRank} did outperform the \textit{Seed} system on the A/B Preference evaluation.

Thresholding at 100 words created lower scored \textit{AlternativeThirds} summaries if only two extracts could not be selected because the word limit would be exceeded (i.e., no final expanded extract). Also, while the top distance-ranked extracts were the substrate for the human summaries, the systems could use a broader range of extracts and create interesting (though less on point) highly rated summaries - e.g. the \textit{Seed} system in (7):

\enspace
\footnotesize
\noindent(7)\textit{If there is such a thing as a hate crime, we saw it at Kroger and we saw it in the synagogue again in Pittsburgh," McConnell said. The Kroger Co. announced today a \$1 million donation to the USO as part of its annual Honoring Our Heroes campaign. Kroger's Honoring Our Heroes campaign has supported veterans, active duty military and their families since 2010, raising more than \$21 million through both corporate funds and customer donations.}

\enspace
\normalsize
While a variety of discourse level extractive summarization approaches attempt to create well-formed discourses, of which specificity and a host of other pragmatic phenomena would follow suit and contribute to higher quality, \textit{sentence ordering} approaches are most similar to what is proposed here. For single documents, maintaining the order of extracts in the source material, has provided positive improvements in quality \cite{104Lin:2002:SMS:1073083.1073160,133McKeown:1999:TMS:315149.315355,84ji-nie-2008-sentence}. Sentence ordering for multi-document summarization is harder as there is no \textit{a priori} structural discourse relationship between documents. Nonetheless, chronology can be inferred and ordered across clusters of documents for improved output \cite{9Barzilay:2002:ISS:1622810.1622812,84ji-nie-2008-sentence,19bollegala-etal-2005-machine,20Bollegala:2010:BAS:1631875.1631910}.

Discourse awareness in our system comes from semantic coherence associated with token distances, and pragmatic (rhetorical) coherence associated with the multi-sentence extractions and the nature of specificity in the extraction sets. Our system is lower complexity compared to other systems, but there is less control over the specific and general nature of the extracts and their ordering. Observed benefits from the extractions within a tightly constrained domain cannot be disregarded. While current research and detection of text specificity (e.g. \citet{Li2017}) shows promise of more control, it remains a very difficult problem.

\section{Conclusion and Future Work}
For short extractive multi-document summaries in the context of our risk mining system, focusing on ordering of information specificity as a means of structuring discourse has provided tangible improvements in output quality. Future experimentation will extend to contexts beyond risk mining to test the generalizability of our method in less controlled environments. Further, as summary thresholds increase, our method may require additional constraints to ensure, for example, that global discourse patterns are adhered to - especially as other non-narrative structures are considered (\textit{see e.g.} \citet{Smith2003ModesOD}). 

As noted, observed improvements do not require intricate control of the extracted information. While greater control to improve output more consistently would certainly be welcome, care must be taken not to overburden the system where it is not clear, based on current research into specificity and discourse, that improvement will be found. Nonetheless, specificity-leaning features improve output in extractive summary discourses in the absence of more in-depth NLP - an encouraging step toward focusing on less well-studied discourse phenomena as a means of progress. 




\bibliography{acl2019}
\bibliographystyle{acl_natbib}

\end{document}